\documentclass[9pt,shortpaper,twoside,web]{ieeecolor}
\usepackage{generic}
\usepackage{cite}
\usepackage{amsmath,amssymb,amsfonts}
\usepackage{algorithmic}
\usepackage{graphicx}
\usepackage{textcomp}
\usepackage{placeins}
\usepackage{caption}
\usepackage{float}

\newcommand{\changed}[1]{\textcolor{black}{#1}}

\def\BibTeX{{\rm B\kern-.05em{\sc i\kern-.025em b}\kern-.08em
    T\kern-.1667em\lower.7ex\hbox{E}\kern-.125emX}}
\begin{document}
\title{Model of the Weak Reset Process in HfOx Resistive Memory \\ for Deep Learning Frameworks}
\author{Atreya Majumdar, Marc Bocquet, Tifenn Hirtzlin, \IEEEmembership{Student Member, IEEE}, Axel Laborieux, \\ Jacques-Olivier Klein, \IEEEmembership{Member, IEEE}, Etienne Nowak, \IEEEmembership{Member, IEEE}, Elisa Vianello, \IEEEmembership{Member, IEEE}, \\ Jean-Michel Portal, \IEEEmembership{Member, IEEE}, and Damien Querlioz, \IEEEmembership{Senior Member, IEEE\vspace{-0.5cm}
}
\thanks{This work was supported in part by the European Research Council Starting Grant NANOINFER under Grant 715872, and in part by the
Agence Nationale de la Recherche grant NEURONIC under Grant ANR-18-CE24-0009.}
\thanks{Atreya Majumdar, Axel Laborieux, Jacques-Olivier Klein, and Damien Querlioz are with Universit\'e Paris-Saclay, CNRS, Centre de Nanosciences et de Nanotechnologies, 91120 Palaiseau, France.}
\thanks{Marc Bocquet, and Jean-Michel Portal are with Institut Matériaux Microélectronique Nanosciences de Provence, Aix Marseille University, Universit\'e de Toulon, 13453 Marseille, France.}
\thanks{Tifenn Hirtzlin, Etienne Nowak, and Elisa Vianello are with CEA, LETI, Universit\'e Grenoble Alpes, Minatec Campus, 38054 Grenoble, France.}}

\maketitle

\begin{abstract}
The implementation of current deep learning training algorithms is power-hungry, owing to data transfer between memory and logic units. Oxide-based RRAMs are outstanding candidates to implement in-memory computing, which is less power-intensive. Their weak RESET regime, is particularly attractive for learning, as it allows tuning the resistance of the devices with remarkable endurance.  However, the resistive change behavior in this regime suffers many fluctuations and is particularly challenging to model, especially in a way compatible with tools used for simulating deep learning. In this work, we present a model of the weak RESET process in hafnium oxide RRAM and integrate this model within the PyTorch deep learning framework. Validated on experiments on a hybrid CMOS/RRAM technology, our model reproduces both the noisy progressive behavior and the device-to-device (D2D) variability.  We use this tool to train Binarized Neural Networks for the MNIST handwritten digit recognition task and the CIFAR-10 object classification task. We simulate our model with and without various aspects of device imperfections to understand their impact on the training process and identify that the D2D variability is the most detrimental aspect. The framework can be used in the same manner for other types of memories to identify the device imperfections that cause the most degradation, which can, in turn, be used to optimize the devices to reduce the impact of these imperfections.
\end{abstract}


\section{Introduction}
\label{sec:introduction}
The  advance of machine learning algorithms holds remarkable prospects in terms of benefits to the society \cite{yu2018neuro}. However, this progress comes at the cost of a considerable energy budget  \cite{markovic2020physics}. The bulk of this energy consumption is attributed to the shuttling of information between the memory and logic units of the computing system \cite{pedram2016dark}, 
a bottleneck, that can be circumvented by the use of
in-memory computing. For such designs, oxide-based resistive memories (RRAMs), or memristors, are a major breakthrough. Their fast, low-power, non-volatile switching and full compatibility with the CMOS process lends quite well towards the realization of energy-efficient, adaptable synaptic weights \cite{grossi2016fundamental, ambrogio2016neuromorphic}. 
Unfortunately, owing to their dependence on nanometer-scale physics of atoms and ions, oxide-based RRAMs are usually very difficult to model accurately, which is a challenge for the design of in-memory neural networks. 

\begin{figure}[!t]
\centerline{\includegraphics[width=1.0\columnwidth]{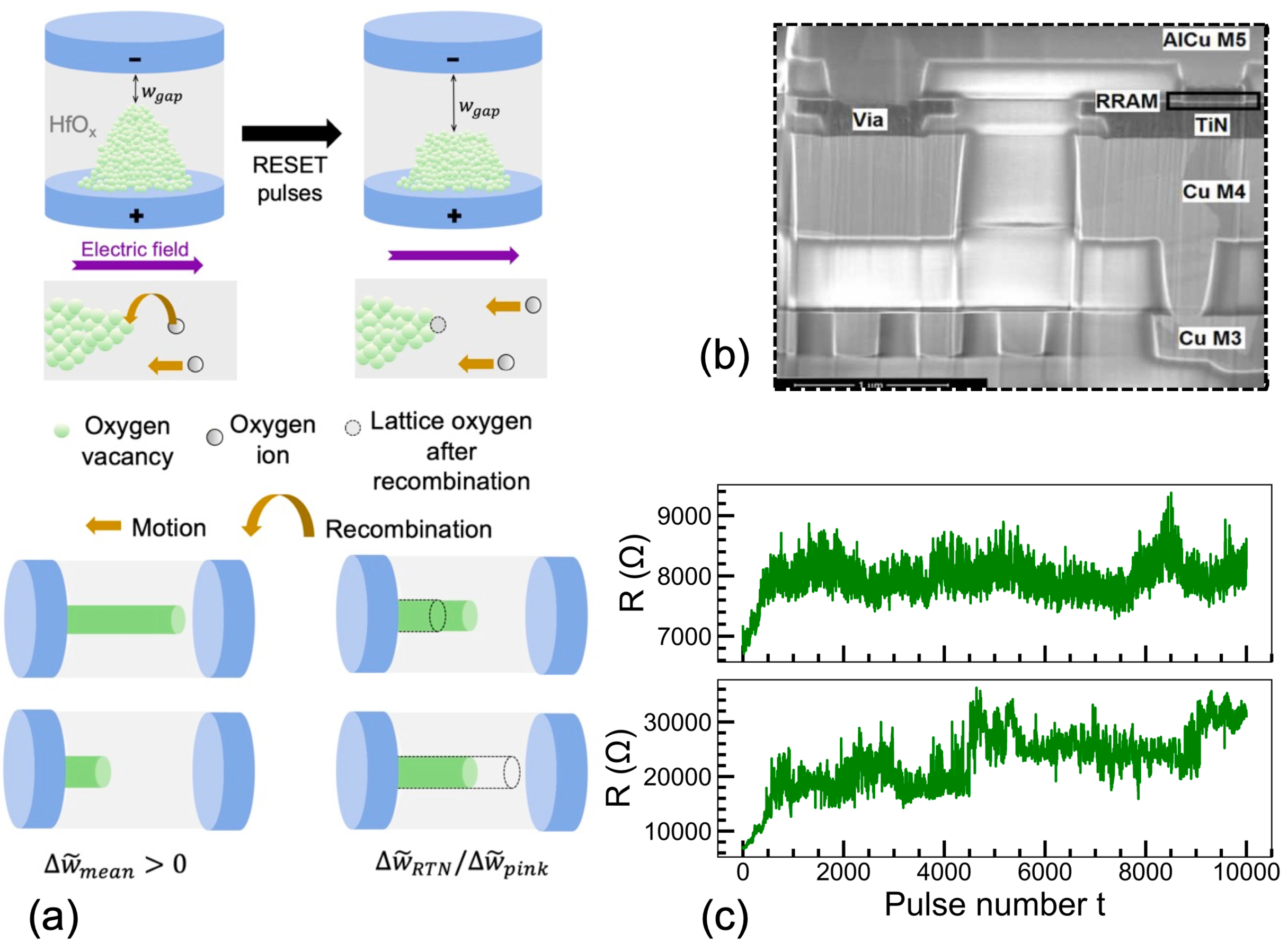}}
\caption{(a) Illustration of the progressive dissolution of the conducting filament by the recombination of oxygen ions and vacancies under the influence of consecutive RESET pulses. \changed{Also, the increment ($\Delta\widetilde{w}_{mean}$) and fluctuation terms ($\Delta\widetilde{w}_{RTN}$/$\Delta\widetilde{w}_{pink}$) of our device model (described in Section~\ref{sec:Device Model}) are shown schematically}. (b) SEM image of an HfOx-based RRAM device integrated into the BEOL of our technology. (c) Progressive evolution of the resistance of two measured  devices with the application of consecutive weak RESET pulses of amplitude 1~V and writing time of 0.1~$\mu$s.}
\vspace{-0.5cm}
\label{fig1}
\end{figure}

Oxide-based RRAM devices switch through the formation and dissolution of conductive filaments of oxygen vacancies (Fig.~\ref{fig1}(a)). They function based on a combination of transport, thermal, and electrochemical effects; a multiplicity of mechanisms of atomic movement can coexist within the same device, giving rise to different regimes, depending on the state of the device and bias conditions \cite{bocquet2014robust}. Additionally, the devices exhibit very important fluctuations that resist simple modeling \cite{ambrogio2014statistical1,  ambrogio2014statistical2}. In recent years, considerable progress has been made in the modeling of these devices in the regimes relevant for embedded and standalone memory applications \cite{bocquet2014robust, ielmini2011modeling, jiang2016compact, li2015variation, chen2018neurosim}. On the other hand, a programming regime known as weak RESET (programmed with a low voltage)\cite{piccolboni2015investigation, hirtzlin2019hybrid} remains vastly unexplored, as this regime, presenting exacerbated fluctuations, has no usage for conventional memory applications. Remarkably, recent works suggest that this regime might be extremely useful for artificial intelligence (AI) and neuromorphic applications, allowing such systems to do learning using little power and area \cite{hirtzlin2019hybrid, zhou2018new}.  \changed{Although studies about low voltage switching\cite{alibart2012high}, device models\cite{kvatinsky2015vteam}, and noise\cite{choi2014random, stathopoulos2019memristive} have been carried out in the past, a comprehensive study integrating all these aspects have not been done.} To investigate this lead convincingly, and to design systems, an accurate model of the weak RESET process is needed. Additionally, the model needs to be compatible with the very specific frameworks used for designing neural networks
(PyTorch, TensorFlow...), 
optimized for operating on graphics processing units (GPUs) and to perform automatic differentiation and were not designed for including device effects such as noise and variability \cite{paszke2017automatic}.

In this work, we propose an efficient analytical  behavioral model for the weak RESET regime of HfO$_\textrm{x}$-based RRAM, including device fluctuations, and implement it within a deep-learning framework to model synaptic parameters. We provide and validate this model with extensive measurements, using multiple statistical quantities,  on a hybrid HfO$_\textrm{x}$ RRAM/CMOS integrated circuit.
This device model is specifically optimized for integration within deep learning frameworks, and this feature allows us to investigate the behavior of such devices in the context of neural network training. We implement this model within PyTorch, a deep-learning framework by adapting the optimizer. We present simulation results of binarized neural network (BNN), a quantized form of more traditional neural networks for which the weak RESET regime of RRAMs is particularly attractive, using fully connected and convolutional architecture for MNIST and CIFAR-10 tasks, respectively. Finally, using these simulations, we study the impact of device imperfections on the network performance.

\section{Hafnium Oxide RRAM Technology}
\label{sec:RRAM Device Characterization}

For this work, we rely on measurements of a hafnium oxide (HfO$_\textrm{x}$)-based OxRAM technology. \changed{The memory stack has a TiN/HfO$_\textrm{x}$(10~nm)/Ti(10~nm)/TiN composition where the TiN layers serve as the electrodes \cite{hirtzlin2020digital}}. Our nanodevices are integrated within the back-end-of-line of a 130~nm commercial CMOS process,  between metal levels four and five, as shown in Fig.~\ref{fig1}(b). Such integration of logic and memory facilitates the implementation of energy-efficient in-memory computing.
Each memory device is associated with an NMOSFET, allowing precise control of the programming conditions, such as the compliance current, which enables the formation of the conducting filaments \cite{ielmini2011modeling}. After an initial electroforming step, the device can switch between low-resistance (LRS) and high-resistance states (HRS) depending upon the polarity of the applied voltage pulses.  The switching between LRS and HRS is attributed to the gradual formation and dissolution of the conductive oxygen-vacancy filaments within the oxide. 

\begin{figure}[!t]
\centerline{\includegraphics[width=\columnwidth]{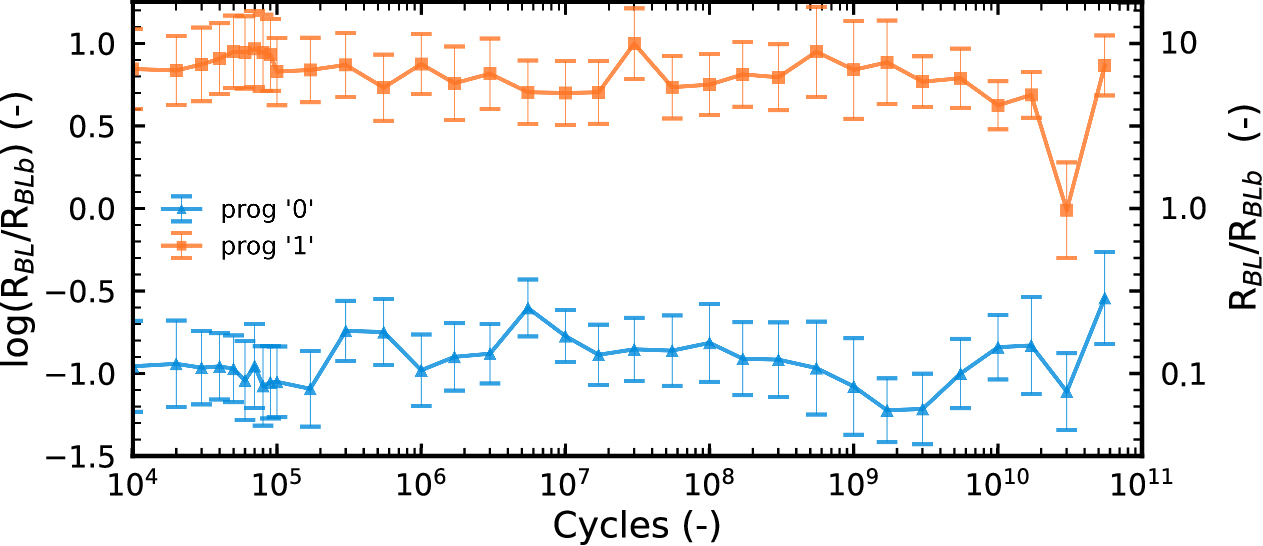}}
\caption{Endurance measurement on two complementary devices programmed with weak RESET pulses of width 1 $\mu$s and SET compliance current of 200 $\mu$A: median value of log resistance ratio ($R_{BL}/R_{BLb}$), extracted over 10k rounds for measurement of a pair of devices over $5\times10^{10}$ cycles.}
\label{fig2}
\end{figure}

The weak RESET regime is stimulated by applying low voltage negative pulses, \changed{and pulse times shorter than used in the traditional RESET}.  \changed{It makes the switching smoother, which enables the finer tuning of resistance, at the cost of a reduced HRS/LRS ratio.} 
Measurements in Fig.~\ref{fig1}(c) show that repeated one-Volt weak RESET pulses lead to a progressive increase in the cell resistance, albeit in a noisy manner. In this Figure, the resistance is read at a very low voltage (0.1~V) after each weak RESET pulse, so that a very low read current flows through the device, and, therefore, there is no read disturb effect.  
We choose the weak RESET regime of operation to achieve high endurance in our devices. For learning tasks, this is essential, as individual devices are required to be programmed reliably for a large number of cycles. \changed{Fig.~\ref{fig2} shows the outstanding endurance of two complementary devices each with resistances R$_{BL}$ and R$_{BLb}$, that are programmed in the weak RESET, for more than $10^{10}$ cycles.} This is orders of magnitude more than when devices are used with traditional higher-voltage RESET \cite{hirtzlin2020digital}, and orders of magnitude more of what is needed for practical learning tasks (e.g., ~$10^4$ cycles for the CIFAR-10 object recognition task).

However, the resistance increase due to the weak RESET seen in Fig.~\ref{fig1}(c) is particularly noisy, and in a way that appears non-trivial. Cells in the weak RESET regime are therefore reminiscent of biological synapses, which also modulate their conductivity (weight) during the learning process, in a way that is often believed to be noisy\cite{rusakov2020noisy}. \changed{Recently, it has been shown that RRAM cells in weak RESET could indeed be used to do learning, for a type of neural networks, called binarized neural networks (BNNs), which are more resilient to noise, and less energy consuming than analog neural networks \cite{hirtzlin2019hybrid, zhou2018new}}. 

\section{Device Characterization and Modeling}
\label{sec:Device Model}

\begin{figure}[t]
\centerline{\includegraphics[width=\columnwidth]{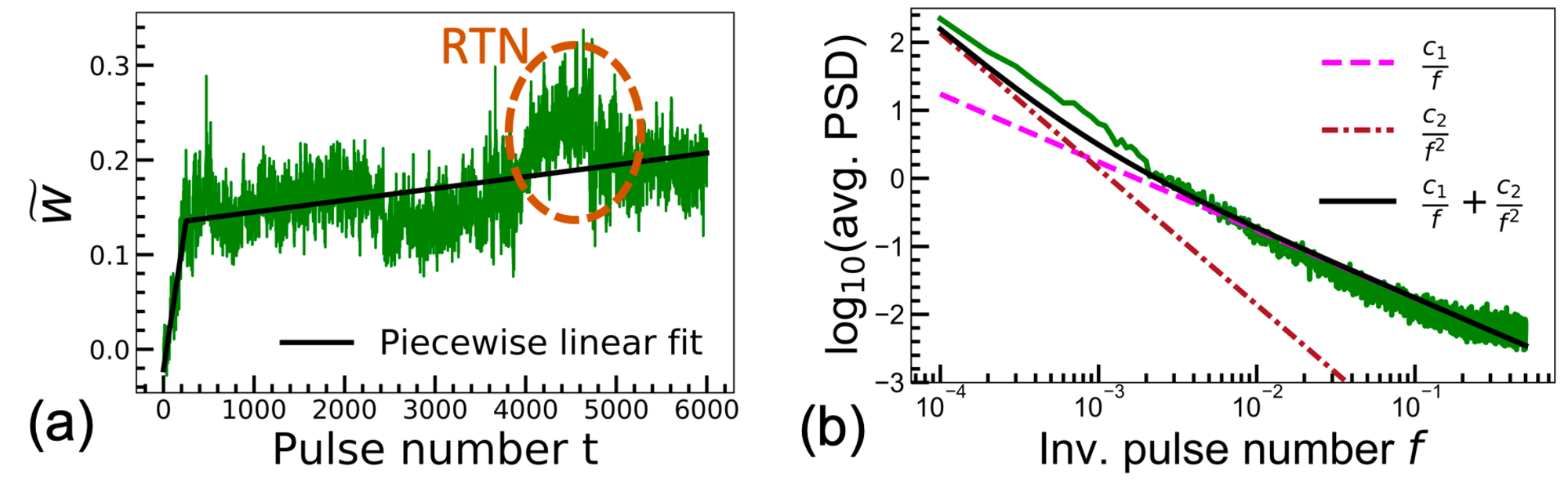}}
\caption{(a) Piecewise linear fit of the mean model to the $\widetilde{w}$ of a device \changed{(different than Fig.~\ref{fig1}(c))}. The parameters $m_1$, $t^*$, $c_1$ and $m_2$ are extracted from this fit. (b) Power spectral density of $\widetilde{w}$ averaged over 64 devices showing the presence of a ($1/f^2$) trend for low frequencies and a pink-noise like ($1/f$) response for higher frequencies.}
\label{fig3}
\end{figure}

In this section, we introduce our device model for the resistance in the weak RESET regime.
To model the weak RESET behavior, we take the established approach of using the tunneling gap between the partially dissolved oxygen-vacancy filament and the electrode (Fig.~\ref{fig1}(a)), $w_{gap}$ as the state parameter \cite{ielmini2011modeling, jiang2016compact}. For practical purpose, we use the dimensionless quantity $\widetilde{w}$,  defined as
\begin{equation}
    \widetilde{w} = {w_{gap}}/{w_0},
    \label{equ1}
\end{equation}
where $w_0$ is a length scale associated with the standard size of the filament. Owing to its quantum mechanical origin, the resistance of the device associated with the tunneling gap $\widetilde{w}$ is 
\begin{equation}
    R(\widetilde{w}) = R_0 \exp(\widetilde{w}),
    \label{equ2}
\end{equation}
where $R_0$ is the resistance of the device in LRS, i.e, when the tunneling gap is zero.  \changed{The model does not include filament diameter, which appears to have a second-order effect during the weak RESET process.}
\changed{The variations in the tunneling gap $\widetilde{w}$ give rise to its progressive RESET behavior. It also leads to the noise which is a consequence of the invasive biasing and is not related to the read noise \cite{raghavan2013microscopic}}. Fig.~\ref{fig3}(a) shows an example of $\widetilde{w}$ extracted from measurements, showing both its increasing trend and its noise, when successive weak RESET pulses are applied. In our model, the earlier contribution, $\widetilde{w}_{mean}$ is described by a piecewise linear model as a function of the pulse number $t$,  parameterized by $m_1$, $c_1$, $t^*$ and $m_2$ as 

\begin{figure*}[t]
    \centering
    \includegraphics[width=0.8\textwidth]{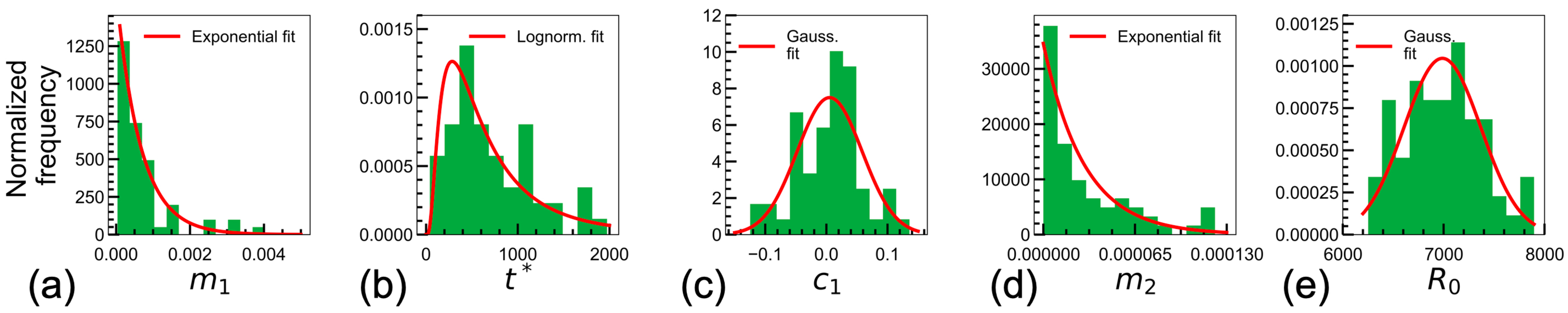}
    \caption{The statistical distribution of the extracted parameters over the 64 devices. The respective slopes for the two regimes $m_1$ and $m_2$ both follow exponential distributions (a) and (d). The threshold pulse number $t^*$ follows a log-normal distribution (b) whereas both the initial intercept $c_1$ (c), and the initial resistance (e) follow Gaussian distributions.}
    \label{fig4}
\end{figure*}
\begin{figure*}[!ht]
\centering
{\includegraphics[width=\textwidth]{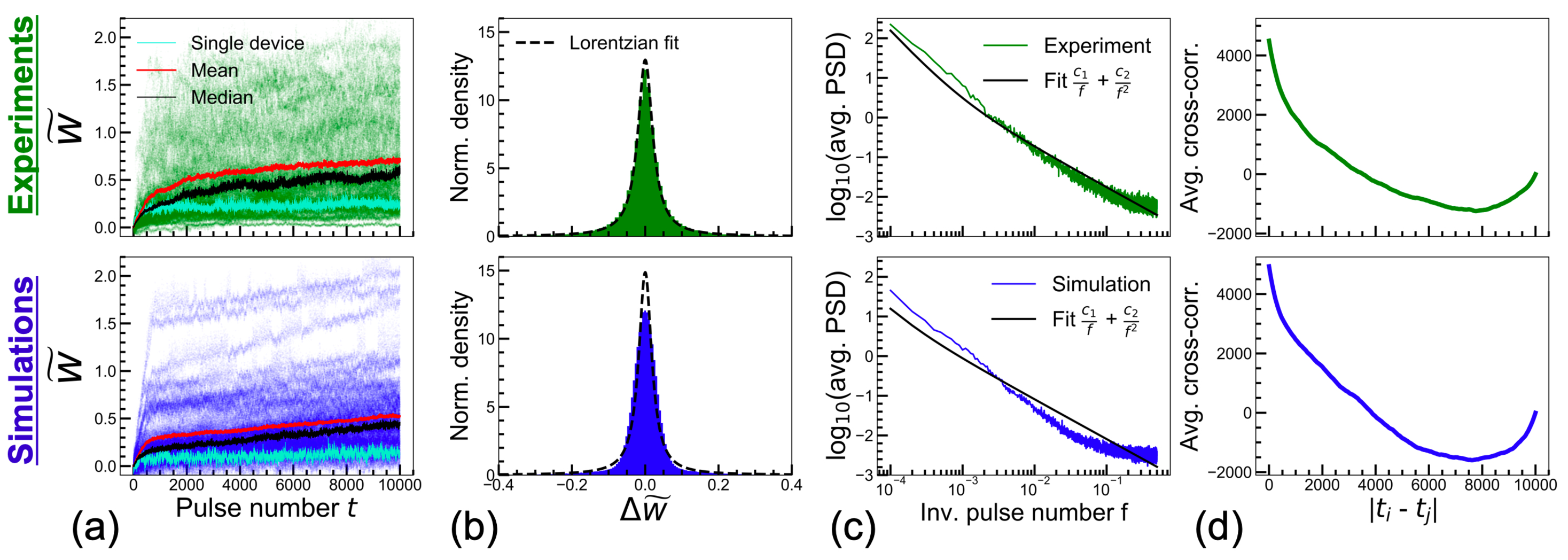}}
\caption{Comparison of experiments and simulations over 64 devices. (a) Scatter plots of $\widetilde{w}$ as a function of the number of applied weak RESET pulses. Also, the evolution of $\widetilde{w}$ for a single device shown for both. (b) Histograms of changes in $\widetilde{w}$ after each pulse. (c) Average power spectral density of $\widetilde{w}$ following Lorentzian distributions. (d) Average cross-correlation between the devices.\vspace{-0.25cm}
}
\label{fig5}
\end{figure*}

\begin{equation}
   \widetilde{w}_{mean} = \left\{
        \begin{array}{ll}
            m_1t+c_1 & t < t^* \\
            m_2t+(m_1-m_2)t^*+c_1 & t \geq t^*.
        \end{array}
    \right.
    \label{equ3}
\end{equation}

The first regime ($t < t^*$), where the increase of resistance is steeper and less noisy is physically related to conditions where the heating due to the Joule effect is more pronounced, compared to the later one ($ t \geq t^*$), where the resistance of the device is higher. Under this condition, the resistance increase is much less monotonic and prone to more noise. 


To characterize the fluctuations in the value of the resistance, we first compute the power spectral density (PSD) of $\widetilde{w}$ extracted from measurements. As shown in Fig.~\ref{fig3}(b), the PSD averaged over 64 devices exhibits both a $1/f^2$ and a $1/f$ contribution. The $1/f^2$ part is consistent with the Random Telegraph Noise (RTN) that we find in our devices (Fig.~\ref{fig3}(a)) \cite{ambrogio2014impact}. On the other hand, the $1/f$ dependence indicates the existence of pink noise. \changed{Both of these types of noise are related to the switching process and is independent of the passive noise that we would get during read-out only.} In our model, we capture these two types of noise by the quantities $\widetilde{w}_{pink}$ and $\widetilde{w}_{RTN}$.

RTN can be found in the second regime of the mean model (Fig.~\ref{fig3}(a)) and is attributed to the perturbations related to the creation and destruction of oxygen vacancies in non-stoichiometric hafnium oxide \cite{raghavan2013microscopic}. The RTN component ($\widetilde{w}_{RTN}$) is modeled as a two-state Markov process  
\begin{equation}
    \widetilde{w}_{RTN} = aX,
    \label{equ4}
\end{equation}
where $X$ is a random variable taking a value of zero or one depending upon the resistive contribution from the fluctuations of the vacancies, and $a$ is the amplitude of the resistance jumps \cite{ito2011modeling}. The probabilities of switching from zero to one and vice-versa are given by $P_{high}$ and $P_{low}$, which are asymmetric. \changed{ Hence, the transition matrix $T$, of the Markov process is defined as $T_{12} = P_{high}$ and $T_{21} = P_{low}$.}


The pink noise, on the other hand, might be related to the dynamically changing defect states in the oxide \cite{grasser2020noise}. It is modeled using an approach where the values can be sequentially generated, which is more suitable for the GPU-based implementation that is expected for deep learning frameworks \cite{kasdin1995discrete}. In this method, the pink noise is generated in the following manner. Firstly, $pole$ number of white Gaussian values, $\omega_r$ ($r=0, 1, ..., pole$) are generated. \changed{These values are then passed through a low-pass FIR filter with coefficients ($b_r$), which are the values of the impulse response, so that the generated noise is pink in its PSD.} Thus, mathematically the pink noise component is described by 
\begin{equation}
    \widetilde{w}_{pink} = \alpha\sum_{r=0}^{pole} b_r\omega_r,
    \label{equ5}
\end{equation}
where $\alpha$ is a scaling factor. 
The variation of our state variable $\widetilde{w}$ is then obtained by the superposition of these three components as  

\begin{equation}
    \widetilde{w} = \widetilde{w}_{mean}+\widetilde{w}_{RTN}+\widetilde{w}_{pink}.
    \label{equ6}
\end{equation}
\changed{The physical impact of the variation of these three terms is shown schematically in Fig.~\ref{fig1}(a)}.

In addition, RRAM devices are subject to important device-to-device (D2D) variability, due to the various possible topologies of the conductive filaments and dynamic perturbations, which can have a considerable impact on neuromorphic applications and should be modeled carefully. Fig.~\ref{fig4}(a)-(d) show the distribution of the parameters of our mean model, extracted from experiments on 64 devices integrated into a memory array. The distributions of the $m_1$, $t^*$, $c_1$ and $m_2$ parameters can be well fitted using an exponential, a lognormal, a Gaussian, and a lognormal distribution respectively.  \changed{The variation in the absolute value of the resistance is done by sampling the initial resistance $R_0$ (Eq.~\ref{equ2}) from a Gaussian distribution whose parameters are extracted from the experimental initial LRS distribution of the devices (Fig.~\ref{fig4}(e)). This makes sure that even if we are only dealing with the variations in the state parameter $\widetilde{w}$, the absolute resistances also bear the same variability as the devices.} Table~\ref{table1} lists the extracted parameters, used for our simulations. \changed{The parameters $m_1$ and $m_2$ describe the monotonic progressive increase of the filament gap, and follow an exponential law, highlighting that some devices are relatively insensitive to weak RESET. The Gaussian distribution of $R_0$ and $c_1$ is connected to the Gaussian distribution of the LRS.}

\changed{The parameters used to generate the noise (cycle-to-cycle variation) are fine-tuned so that the experiments and simulations in Fig.\ref{fig5}(b) and (c) match, and the values are summarized in Table~\ref{table2}. The obtained values naturally replicate the noise levels observed in both regimes of the mean model.} 

Fig.~\ref{fig5} shows that the resulting model, integrating D2D, reproduces all measured aspects of the experiments with outstanding accuracy. Fig.~\ref{fig5}(a) shows the individual trajectories in the weak RESET process of 64 measured and 64 simulated devices. Fig.~\ref{fig5}(b) shows that the distribution of the changes in $\widetilde{w}$ after each weak RESET pulse follows the same Lorentzian distribution in both the experiments and simulations. \changed{It is centered at zero, which implies that the fluctuations dominate over the monotonic changes arising from the mean model, which would have caused a bimodal distribution of positive values.} The narrow peaks and wide tails of Lorentzian distribution represent the more frequent pink-noise and the less frequent RTN induced fluctuations respectively. Fig.~\ref{fig5}(c) shows the mean spectral power spectrum of Fig.~\ref{fig5}(a), \changed{and Fig.~\ref{fig5}(d) shows the mean cross-correlation of $\widetilde{w}$ between the 64 devices, where $t_i$ and $t_j$ are the pulse numbers to the $i^\textrm{th}$ and $j^\textrm{th}$ devices, defined as:
\begin{equation}
    Cross\:Corr.\:(|t_i-t_j|)=\sum_{all\:t_i,t_j}\widetilde{w}(t_i)\widetilde{w}(t_j).
\end{equation}} 
The average cross-correlation between the 64 devices is a measure of the D2D variability captured by our model, which also agrees with the experiments. \changed{The mean auto-correlation at zero shift is about 10,000, which is twice the average cross-correlation at zero shift, indicating that the inter-device variability is larger than the intra-device one. Overall,} the model, therefore, seems ideal to mimic RRAM cells.

\begin{table}[t]
\centering
\caption{\textbf{Device-to-device variation}: Parameters for RTN, mean model components and the initial resistances that characterizes the variability between devices and extracted from Fig.~\ref{fig4}. The probability density functions had the following forms:\\ 
$f_{uniform}(x; a_1, a_2)=\frac{1}{a_2-a_1}$, $a_1 \leq x \leq a_2$.\\ $f_{exponential}(x; x_0, \lambda)=\frac{1}{\lambda}\exp\big({-\frac{x-x_0}{\lambda}}\big),  x \geq x_0 $.\\ 
$f_{Gaussian}(x; \mu, \sigma)=\frac{1}{\sigma\sqrt{2\pi}}\exp{\Big[-\frac{1}{2}\Big(\frac{x-\mu}{\sigma}\Big)^2\Big]}$.\\
$f_{lognormal}(x; s, \sigma)= \frac{1}{sx\sqrt{2\pi}}\exp{\Big[-\frac{1}{2}\Big(\frac{log(x/\sigma)}{s}\Big)^2\Big]}, x \geq 0 $.}
\begin{tabular}{|c|c|c|c|}
\hline
Component & Model param. & Distr. & Distr. param. \\ \hline
{RTN amplitude}& $a$ \rule{0pt}{2.2ex}& Uniform & $a_1$=0, $a_2$=0.5 \\ \hline
& $m_1$ \rule{0pt}{2.2ex} & Exponential & $x_0$=3.74e-5,\\
& & & $\lambda$=6.56e-4  \\ \cline{2-4} 
{Mean model} & $c_1$ \rule{0pt}{2.2ex}& Gaussian & $\mu$=5.29e-3,\\
& & & $\sigma$=5.32e-2  \\ \cline{2-4} 
& $t^{*}$ \rule{0pt}{2.2ex}& Log-normal & $s$=0.80,\\
& & & $\sigma$=542.5  \\ \cline{2-4}
& $m_2$ \rule{0pt}{2.2ex}& Exponential & $x_0$=1.64e-34,\\
& & & $\lambda$=2.89e-5  \\ \hline 
{Resistance} \rule{0pt}{2.2ex} & $R_0$& Gaussian & $\mu$=6988~$\Omega$,\\
& & & $\sigma$=381.7~$\Omega$  \\ \hline
\end{tabular}
\vspace{-0.5cm}
\label{table1}
\end{table}

\begin{table}[t]
\centering
\caption{\textbf{Cycle-to-cycle variation}: Parameters for RTN and pink noise that account for the noise on the resistance of devices.}
\begin{tabular}{|c|c|c|}
\hline
Component & Model param. & Value \\ \hline
{RTN} & $P_{high}$ \rule{0pt}{2ex}& 0.0008 \\\cline{2-3}
& $P_{low}$ \rule{0pt}{2.2ex}   & 0.002 \\ \hline
{Pink noise} \rule{0pt}{2.2ex} & $\alpha$  & 0.025 \\\cline{2-3} 
& $pole$ \rule{0pt}{2.2ex}& 15\\ \hline
\end{tabular}
\label{table2}
\vspace{-0.25cm}
\end{table}

\begin{figure}[t]
\centerline{\includegraphics[width=0.9\columnwidth]{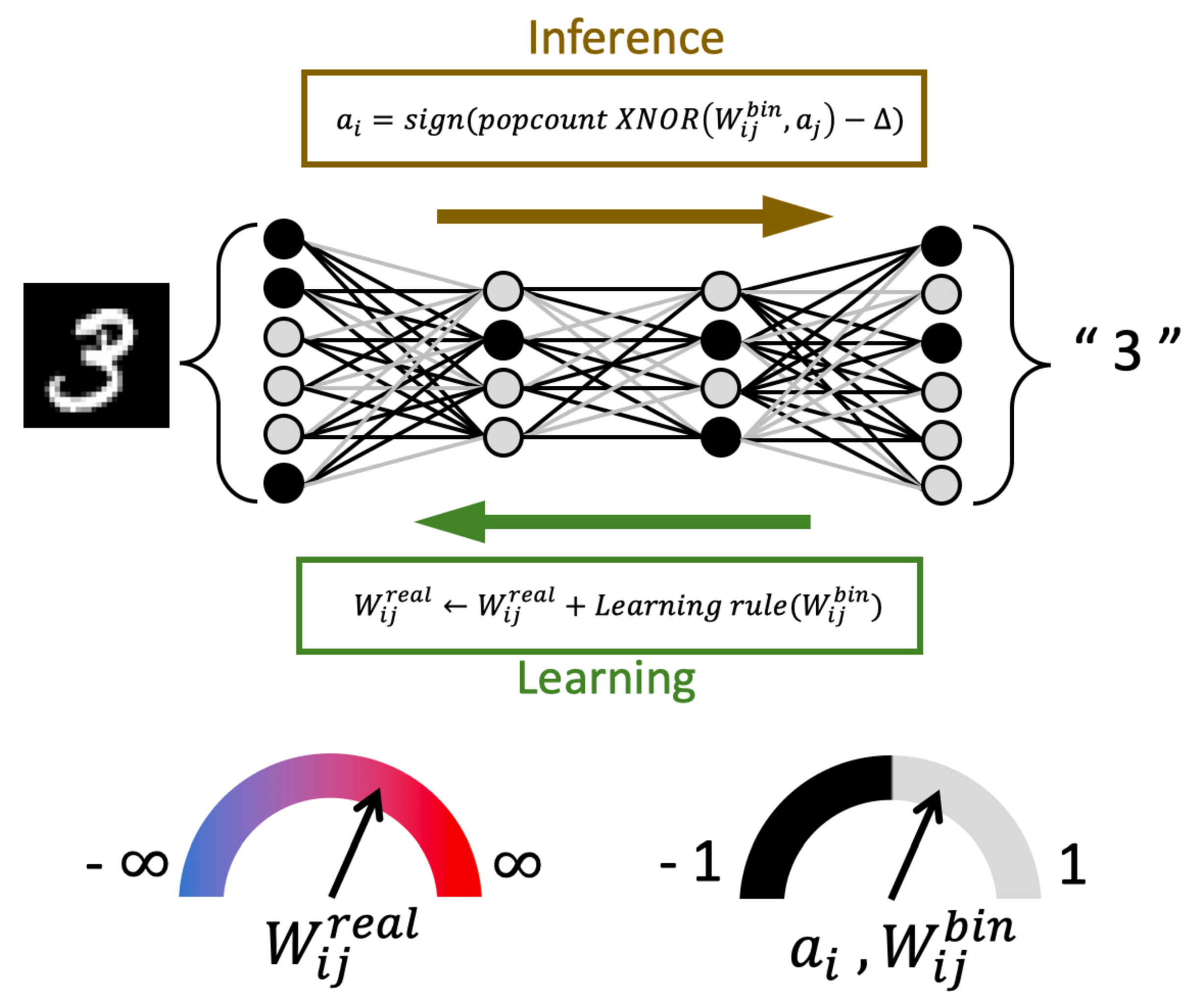}}
\caption{Principle of operation of a Binarized Neural Network (BNN) showing the forward pass (inference) and backward pass (learning or training). The inference depends only on the binarized weights $W_{ij}^{bin}$, whereas the training involves updating the real weights $W_{ij}^{real}$.}
\label{fig6}
\end{figure}

\section{Implementation within a Deep Learning Framework}
\label{sec:Implementation within a Deep Learning Framework}

Artificial neural networks (ANN) are  networks of neurons, connected by synapses, laid in a hierarchical manner: the neuronal activations of a layer are computed from the neurons of the previous layer. The value of neuron activation is computed by first taking the sum over the previous activations weighted by their corresponding synaptic values, and then, applying a non-linear function  to it. The aim of learning a task is to find an optimum set of values for the synaptic connections, called weights. To that end, analog memory cells have been used as the weights owing to their ability to adapt conductances \cite{ambrogio2018equivalent}.

However, to train ANNs, precise values of these weights need to be stored and updated, since the weights and activations can take any real value. This is a problem for RRAM-based implementation in the weak RESET regime, as inter-device and intra-device variabilities are ubiquitous in such nano-devices, as seen in section~\ref{sec:Device Model}. An alternative approach is to use BNNs, where both the neuronal activations and synaptic weights take binary values (+1 and -1) \cite{hubara2016binarized, rastegari2016xnor}. Despite this simplicity of representation, BNNs can approach state-of-the-art accuracy on vision tasks \cite{hubara2016binarized}. During inference, that is calculating the output of the network given the input, this makes their arithmetic extremely simple. The product of the activation and the weight is replaced by a simple XNOR operation. Also, the accumulation of the products can be simply done by counting the number of ones, called the population count. Both of these can be implemented using relatively simple, low power consuming circuitry \cite{hirtzlin2020digital}.  \changed{The advantage of binarization is both from the reduction of the read-out complexity as well as from the fact that low precision synapses and weights can be used for inference.} 

\begin{figure}[t]
\centerline{\includegraphics[width=\columnwidth]{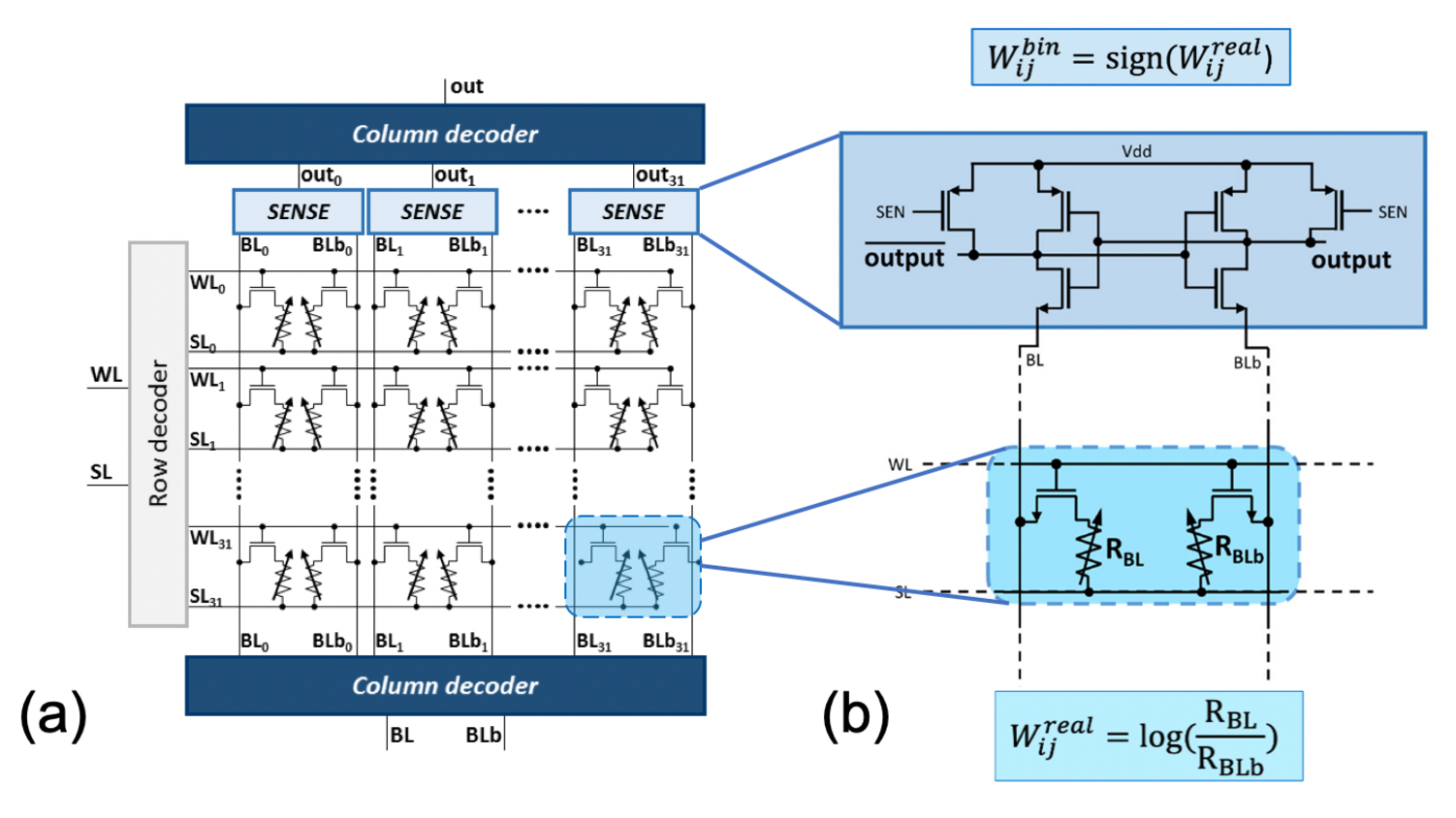}}
\caption{(a) Schematic of the 2T2R memory array used for implementing BNNs. (b) Circuit of the sense amplifier used to extract the binary weight from a 2T2R synapse, along with equations showing how the resistances $R_{BL}$ and $R_{BLb}$ connect with the real ($W_{ij}^{real}$) and binary ($W_{ij}^{bin}$) weights.}
\label{fig7}
\end{figure}

During the training phase i.e., when the network learns the optimum values of the weights, a hidden, real-valued weight is also associated with the synapses \cite{hubara2016binarized, rastegari2016xnor}. 
As shown in Fig.~\ref{fig6}, the binarized weight $W_{ij}^{bin}$ connecting the neuron $a_j$ of the previous layer to the neuron $a_i$ of the next layer relates to the hidden real weight, $W_{ij}^{real}$, as 
\begin{equation}
    W_{ij}^{bin} = sign(W_{ij}^{real}),
    \label{equ7}
\end{equation}
and the binarized activation is given by
\begin{equation}
    a_i = sign(popcount XNOR(W_{ij}^{bin},a_j) - \Delta),
    \label{equ8}
\end{equation}
where $\Delta$ is a threshold that serves the role of shifting in batch normalization of the activation values. 
During the inference phase, only the binarized weights need to be calculated. On the other hand, for learning, the hidden real weights need to be updated by a learning rule, but not explicitly read. We utilize this by avoiding the use of energy-intensive circuits that are required to read the analog resistance state that plays the role of the real weights. Following the approach of \cite{hirtzlin2019hybrid}, we employ a  differential 2T2R structure within a crossbar array (Fig.~\ref{fig7}(a)), in which the two resistances $R_{BL}$ and $R_{BLb}$ account for a single real synaptic weight as  
\begin{equation}
    W_{ij}^{real} = \log_{10}  \left( R_{BL}/R_{BLb} \right).
    \label{equ9}
\end{equation}
\changed{As shown in \cite{bocquet2018memory} and \cite{lastras2021ratio}, the 2T2R scheme based on the ratio of two resistances provides a lower error rate compared to 1T1R which is crucial for the device to operate in the weak RESET regime. The 2T2R structure also allows performing training relying solely on RESET pulses.}
 
In the training phase, to update the real weight, the RRAM devices are programmed using weak RESET pulses on either of the two devices.
If the BNN learning rule suggests to increase the real weight by $\delta W_{ij}^{real}$, we apply weak RESET pulses to the BL device, therefore increasing $W_{ij}^{real}$. Conversely, if $\delta W_{ij}^{real}$ is negative, we apply weak RESET pulses to the BLb device, therefore reducing, $W_{ij}^{real}$.
In both cases, the number of pulses is chosen proportionally to  $\delta W_{ij}^{real}$.
Due to the differential 2T2R nature of the synapses, this training technique requires only RESET pulses. \changed{For the tasks we have performed, we have seen that the progressivity of the RESET process is sufficient, however for more complex tasks this might not be enough. In that case, we can apply a reprogramming strategy, proposed in \cite{hirtzlin2019hybrid}, to bring back the system where proper RESET is applicable.}

For the inference, the sign of this real hidden weight has to be read, and this can be achieved by an energy-efficient and fast circuit called pre-charge sense amplifier \cite{zhao2014synchronous,hirtzlin2020digital}. It compares $R_{BL}$ and $R_{BLb}$ to give an output of +1 when the former is larger and -1 for the opposite. Fig.~\ref{fig7}(b) shows how the real and binarized weights are computed in the circuit.

\begin{figure}[!t]
\centerline{\includegraphics[width=\columnwidth]{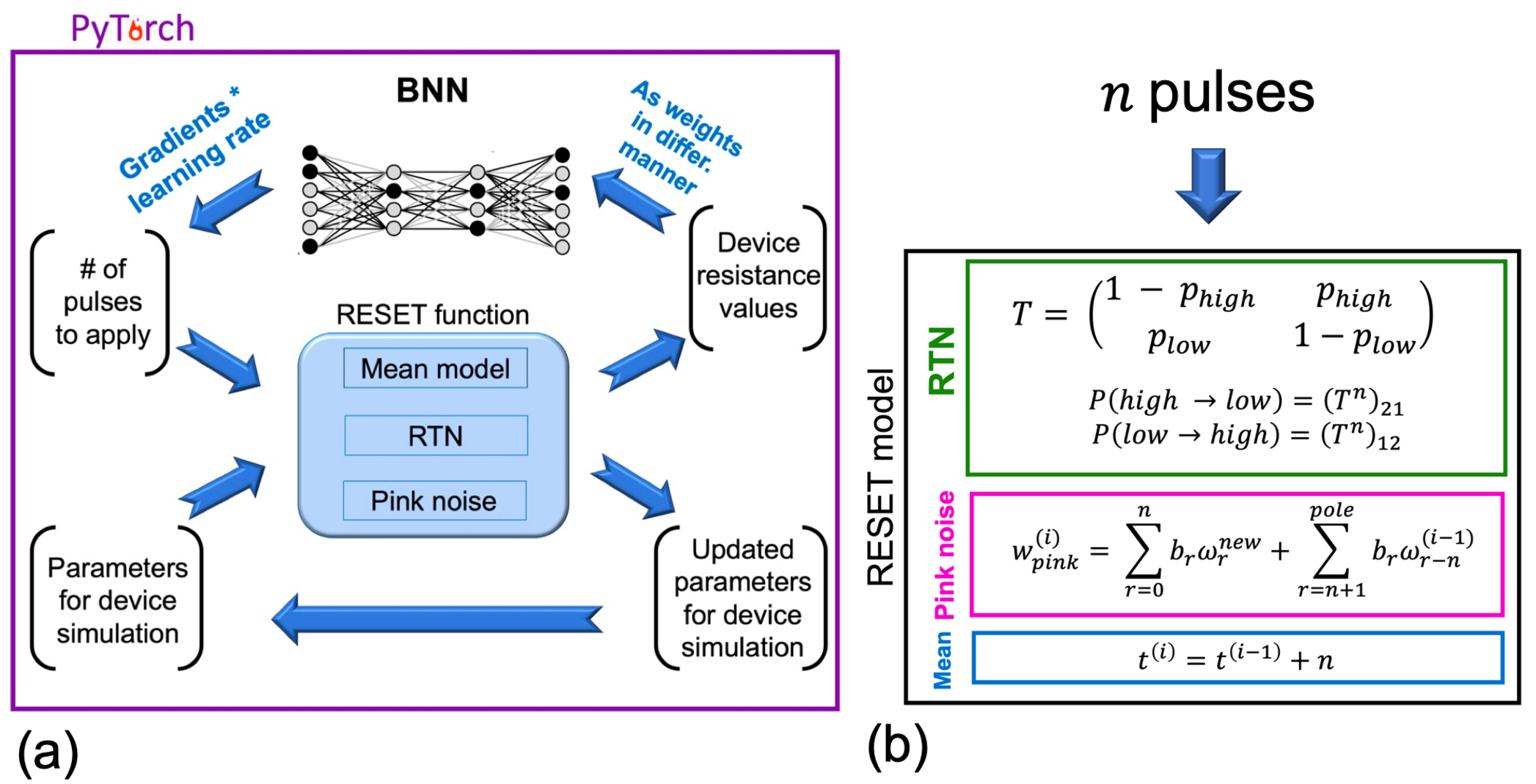}}
\caption{(a) Integration of device simulation and neural network learning within the PyTorch framework. The device resistances act as the synaptic weights in a differential manner, and they are updated according to the device model from the updates provided by the network in the backward pass. (b) The equations for the mean, RTN, and pink noise components inside the RESET function that models the programming of RRAMs within the PyTorch adaptive moment estimation optimizer.
\vspace{-0.5cm}
}
\label{fig8}
\end{figure}

\begin{figure*}[!ht]
\centerline{\includegraphics[width=0.75\textwidth]{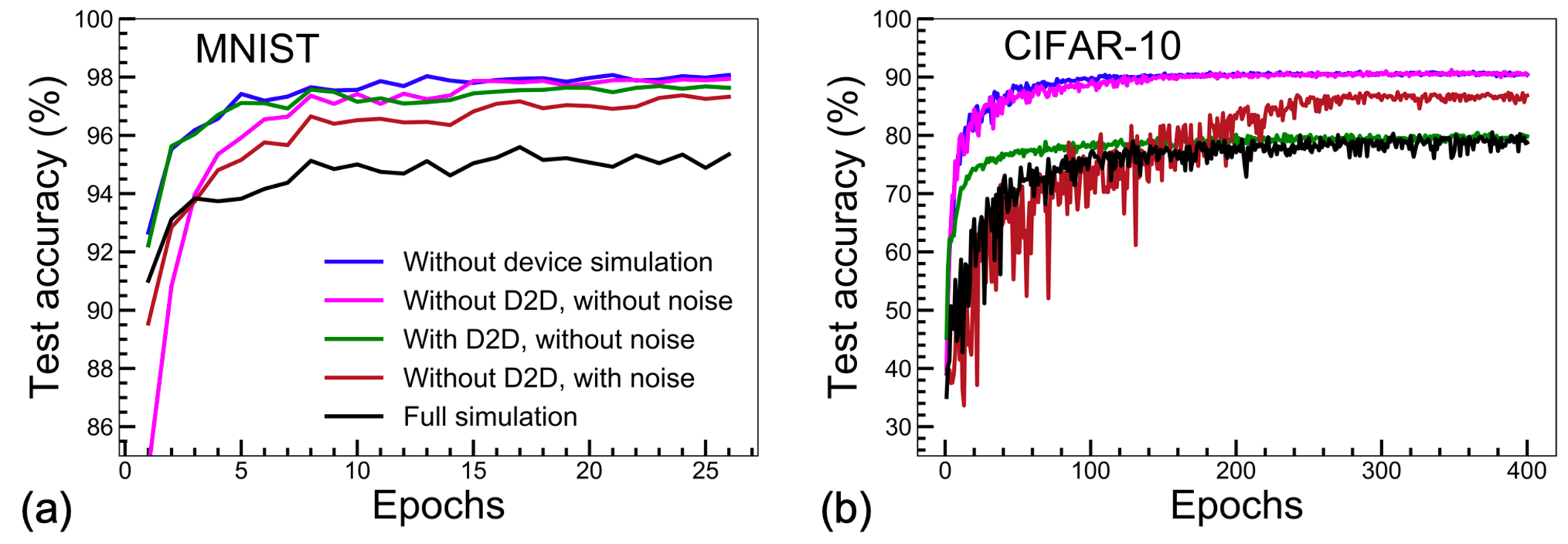}}
\caption{Impact of noise and device-to-device variability on the performances of the binarized neural networks for the (a) MNIST and (b) CIFAR-10 tasks. The plots show the test accuracy during training for five different cases - without device simulation (blue), without both D2D variation and noise (pink), with D2D variation but without noise (green), without D2D but with the full noise and mean model simulation (brown), and the full simulation incorporating both the D2D variability, mean model and noise (black).}
\label{fig9}
\end{figure*}

The frameworks normally used for designing neural networks, such as PyTorch and TensorFlow, model synapses as floating-point real weights. When a neural network is trained, sophisticated optimization algorithms, called optimizers, such as adaptive moment estimation, optimize these weights values by making noiseless, highly precise, and deterministic updates \cite{kingma2015adam}. To test our vision, i.e., to design a physical model where synapses are implemented by RRAM, and the weights are updated using weak RESETs, we adapted the PyTorch deep learning framework in three important ways. 
    First, in deep learning frameworks, the synaptic parameters are stored as tensors with dimensions appropriate to the corresponding architecture. In our approach, these parameters are now modeled by an added dimension that accounts for the device state variables. These are the different parameters that are needed to store the number of pulses that have been previously applied to a device, and to generate pink and telegraph noise (pulses already applied $t$, RTN state variable $X$ and $\omega_r$s).
    
    Second, the parameters of the neural network is typically initialized according to certain pre-defined initialization schemes \cite{he2015delving, lecun2012efficient}. In our case, as the synaptic parameters are linked with the device resistances, we initialize the devices by sampling through the distributions mentioned in Table~\ref{table1}.
    
    Finally, the in-built optimizers provide updates that are real-valued floating-point numbers. But, in RRAM-based networks, we can only modify the resistances, by the application of a discrete number of voltage pulses. Thus, the updates given by PyTorch's adaptive moment estimation methods are discretized by the multiplication by a suitable learning rate and rounding down to integer values. These pulses then produce the synaptic updates following the model of section~\ref{sec:Device Model}. 

The scheme of integration of our device model into the PyTorch framework is schematically shown in Fig.~\ref{fig8}(a). Synaptic weights are initialized as device resistance values in a differential manner incorporating the D2D variability explored in section~\ref{sec:Device Model}. The network does the forward pass on the input and calculates the updates for the weights, which is then converted to integer-valued pulses numbers $n$ that are to be applied to the devices. Using the number of pulses, and the device-based parameters the new device resistance states are calculated as shown in Fig.~\ref{fig8}(b). 

The RTN, pink noise, and mean model components are calculated separately. The RTN state variable is calculated from exponentiating the transition matrix $T$ to the $n^{\textrm{th}}$ power. Pink noise values are generated by drawing $n$ new Gaussian white random numbers and combining them with the already existing $(pole-n)$ values. And, for the mean model component, the pulse number $n$ is simply added to the number of pulses already applied $t$. Now, with the new device resistances, and equivalently the new synaptic weights, the network continues onto the next forward pass.

\section{Neural Network Simulation Results}

We now test our device model, integrated into PyTorch on two pattern recognition tasks. \changed{First, we train a fully connected (FC) BNN with one hidden layer of 3,000 units for solving the MNIST handwritten digit-recognition benchmark. We then train a convolutional BNN to solve the CIFAR-10 object-recognition task. 
The architecture uses 3x3 kernels for convolutions (Conv), and 2x2 for MaxPool (MP) and reads: [Conv384, Conv384, MP, Conv768, Conv768, MP, Conv1536, Conv512, MP, FC(1024-1024-10)].} Figs.~\ref{fig9}(a) and (b) show PyTorch simulations of the training process of binarized neural network for the MNIST and CIFAR-10 tasks, respectively. Test accuracies of 98\% and 90\%  on the MNIST and CIFAR-10 tasks, respectively, were achieved without device simulations (ideal floating-point synapses). 
\changed{Inclusion of the full device simulation in the BNN training simulation makes it four times slower, the bottleneck being the sequential generation of pink noise.}

Incorporating the RRAM model allows testing how various aspects of the RRAM imperfections affect the training performance. We first performed simulations, including the device model, but where the noise and the D2D variability were artificially deactivated (see Figs.~\ref{fig9}(a)-(b)). We observe that for both  tasks, the network can reach the baseline accuracy. Thus, our BNN scheme is  robust to the non-linearity of the devices, which is a major advantage with regards to non-binarized techniques \cite{hirtzlin2019hybrid}. Also, this result highlights that the conversion of floating-point updates to a discrete number of pulses had little effect on the final accuracy.

Figs.~\ref{fig9}(a) and (b) also show that, upon the introduction of noise (both RTN and pink) only,  a point accuracy degradation of 1\% and 2.5\% for the MNIST and CIFAR-10 tasks is obtained. 
Adding D2D variability, the respective degradation of point accuracies are 3\% and 10\%. Also, to identify the impact of the noise independently, we performed simulations with only the noise components artificially deactivated. For MNIST, we find a point degradation of 0.3\% whereas, for the CIFAR-10, it is 10\%. 
For both tasks, the inclusion of the D2D variability therefore caused degradation of test accuracy, although it is more prominent in the CIFAR-10 task (Figs.\ref{fig9}(a)-(b)). 

These results highlight that neural networks have the potential to fully benefit from the advantageous properties of weak RESET (progressivity, high endurance), without suffering from its high level of fluctuations. Also, via this kind of modeling we explored the effects of D2D variability, noise, and non-linearity in greater detail than is possible with only experimental studies. 

\vspace{-0.25cm}
\section{Conclusion}
\label{sec:Conclusion}
In this work, we presented a model of the weak RESET behavior of HfO$_\textrm{x}$ RRAM and its fluctuations, and its integration within a deep learning framework for simulations of hardware neural networks on GPUs. The results suggest the outstanding potential of the weak RESET regime in such conditions. This work also explores the various aspects of RRAM device imperfections on neural network performance. 
Using the proposed framework, future work will investigate the design of more advanced neural networks on difficult tasks, and how neural network design can be optimized for robustness to the fluctuations of RRAM technology. Our modified PyTorch optimizer could also be adapted to all kind of emerging devices considered for neuromorphic applications.  
\vspace{-0.25cm}










\begin{thebibliography}{10}
\providecommand{\url}[1]{#1}
\csname url@samestyle\endcsname
\providecommand{\newblock}{\relax}
\providecommand{\bibinfo}[2]{#2}
\providecommand{\BIBentrySTDinterwordspacing}{\spaceskip=0pt\relax}
\providecommand{\BIBentryALTinterwordstretchfactor}{4}
\providecommand{\BIBentryALTinterwordspacing}{\spaceskip=\fontdimen2\font plus
\BIBentryALTinterwordstretchfactor\fontdimen3\font minus
  \fontdimen4\font\relax}
\providecommand{\BIBforeignlanguage}[2]{{%
\expandafter\ifx\csname l@#1\endcsname\relax
\typeout{** WARNING: IEEEtran.bst: No hyphenation pattern has been}%
\typeout{** loaded for the language `#1'. Using the pattern for}%
\typeout{** the default language instead.}%
\else
\language=\csname l@#1\endcsname
\fi
#2}}
\providecommand{\BIBdecl}{\relax}
\BIBdecl

\bibitem{yu2018neuro}
\BIBentryALTinterwordspacing
S.~Yu, ``Neuro-inspired computing with emerging nonvolatile memorys,''
  \emph{Proc. IEEE}, vol. 106, no.~2, pp. 260--285, 2018. [Online]. Available:
  \url{10.1109/JPROC.2018.2790840}
\BIBentrySTDinterwordspacing

\bibitem{markovic2020physics}
\BIBentryALTinterwordspacing
D.~Markovi{\'c}, A.~Mizrahi, D.~Querlioz, and J.~Grollier, ``Physics for
  neuromorphic computing,'' \emph{Nature Reviews Physics}, vol.~2, no.~9, pp.
  499--510, 2020. [Online]. Available: \url{10.1038/s42254-020-0208-2}
\BIBentrySTDinterwordspacing

\bibitem{pedram2016dark}
\BIBentryALTinterwordspacing
A.~Pedram, S.~Richardson, M.~Horowitz, S.~Galal, and S.~Kvatinsky, ``Dark
  memory and accelerator-rich system optimization in the dark silicon era,''
  \emph{IEEE Des. Test}, vol.~34, no.~2, pp. 39--50, 2016. [Online]. Available:
  \url{10.1109/MDAT.2016.2573586}
\BIBentrySTDinterwordspacing

\bibitem{grossi2016fundamental}
\BIBentryALTinterwordspacing
A.~Grossi, E.~Nowak, C.~Zambelli, C.~Pellissier, S.~Bernasconi, G.~Cibrario,
  K.~El~Hajjam, R.~Crochemore, J.~Nodin, P.~Olivo, and L.~Perniola,
  ``Fundamental variability limits of filament-based rram,'' in \emph{IEDM
  Tech. Dig.}\hskip 1em plus 0.5em minus 0.4em\relax IEEE, 2016, pp. 4--7.
  [Online]. Available: \url{10.1109/IEDM.2016.7838348}
\BIBentrySTDinterwordspacing

\bibitem{ambrogio2016neuromorphic}
\BIBentryALTinterwordspacing
S.~Ambrogio, S.~Balatti, V.~Milo, R.~Carboni, Z.-Q. Wang, A.~Calderoni,
  N.~Ramaswamy, and D.~Ielmini, ``Neuromorphic learning and recognition with
  one-transistor-one-resistor synapses and bistable metal oxide rram,''
  \emph{{IEEE} Trans. Electron Devices}, vol.~63, no.~4, pp. 1508--1515, 2016.
  [Online]. Available: \url{10.1109/TED.2016.2526647}
\BIBentrySTDinterwordspacing

\bibitem{bocquet2014robust}
\BIBentryALTinterwordspacing
M.~Bocquet, D.~Deleruyelle, H.~Aziza, C.~Muller, J.-M. Portal, T.~Cabout, and
  E.~Jalaguier, ``Robust compact model for bipolar oxide-based resistive
  switching memories,'' \emph{{IEEE} Trans. Electron Devices}, vol.~61, no.~3,
  pp. 674--681, 2014. [Online]. Available: \url{10.1109/TED.2013.2296793}
\BIBentrySTDinterwordspacing

\bibitem{ambrogio2014statistical1}
\BIBentryALTinterwordspacing
S.~Ambrogio, S.~Balatti, A.~Cubeta, A.~Calderoni, N.~Ramaswamy, and D.~Ielmini,
  ``Statistical fluctuations in hfo x resistive-switching memory: part
  i-set/reset variability,'' \emph{{IEEE} Trans. Electron Devices}, vol.~61,
  no.~8, pp. 2912--2919, 2014. [Online]. Available:
  \url{10.1109/TED.2014.2330200}
\BIBentrySTDinterwordspacing

\bibitem{ambrogio2014statistical2}
\BIBentryALTinterwordspacing
------, ``Statistical fluctuations in hfo x resistive-switching memory: Part
  ii—random telegraph noise,'' \emph{{IEEE} Trans. Electron Devices},
  vol.~61, no.~8, pp. 2920--2927, 2014. [Online]. Available:
  \url{10.1109/TED.2014.2330202}
\BIBentrySTDinterwordspacing

\bibitem{ielmini2011modeling}
\BIBentryALTinterwordspacing
D.~Ielmini, ``Modeling the universal set/reset characteristics of bipolar rram
  by field-and temperature-driven filament growth,'' \emph{{IEEE} Trans.
  Electron Devices}, vol.~58, no.~12, pp. 4309--4317, 2011. [Online].
  Available: \url{10.1109/TED.2011.2167513}
\BIBentrySTDinterwordspacing

\bibitem{jiang2016compact}
\BIBentryALTinterwordspacing
Z.~Jiang, Y.~Wu, S.~Yu, L.~Yang, K.~Song, Z.~Karim, and H.-S.~P. Wong, ``A
  compact model for metal--oxide resistive random access memory with experiment
  verification,'' \emph{{IEEE} Trans. Electron Devices}, vol.~63, no.~5, pp.
  1884--1892, 2016. [Online]. Available: \url{10.1109/TED.2016.2545412}
\BIBentrySTDinterwordspacing

\bibitem{li2015variation}
\BIBentryALTinterwordspacing
H.~Li, Z.~Jiang, P.~Huang, Y.~Wu, H.-Y. Chen, B.~Gao, X.~Liu, J.~Kang, and
  H.-S. Wong, ``Variation-aware, reliability-emphasized design and optimization
  of rram using spice model,'' in \emph{Proc. DATE}.\hskip 1em plus 0.5em minus
  0.4em\relax IEEE, 2015, pp. 1425--1430. [Online]. Available:
  \url{10.7873/DATE.2015.0362}
\BIBentrySTDinterwordspacing

\bibitem{chen2018neurosim}
\BIBentryALTinterwordspacing
P.-Y. Chen, X.~Peng, and S.~Yu, ``Neurosim: A circuit-level macro model for
  benchmarking neuro-inspired architectures in online learning,'' \emph{{IEEE}
  Trans. Comput.-Aided Design Integr. Circuits Syst.}, vol.~37, no.~12, pp.
  3067--3080, 2018. [Online]. Available: \url{10.1109/TCAD.2018.2789723}
\BIBentrySTDinterwordspacing

\bibitem{piccolboni2015investigation}
\BIBentryALTinterwordspacing
G.~Piccolboni, G.~Molas, J.~M. Portal, R.~Coquand, M.~Bocquet, D.~Garbin,
  E.~Vianello, C.~Carabasse, V.~Delaye, C.~Pellissier, T.~Magis, C.~Cagli,
  M.~Gely, O.~Cueto, D.~Deleruyelle, G.~Ghibaudo, B.~De~Salvo, and L.~Perniola,
  ``Investigation of the potentialities of vertical resistive ram (vrram) for
  neuromorphic applications,'' in \emph{IEDM Tech. Dig.}\hskip 1em plus 0.5em
  minus 0.4em\relax IEEE, 2015, pp. 17--2. [Online]. Available:
  \url{10.1109/IEDM.2015.7409717}
\BIBentrySTDinterwordspacing

\bibitem{hirtzlin2019hybrid}
\BIBentryALTinterwordspacing
T.~Hirtzlin, M.~Bocquet, M.~Ernoult, J.-O. Klein, E.~Nowak, E.~Vianello, J.-M.
  Portal, and D.~Querlioz, ``Hybrid analog-digital learning with differential
  rram synapses,'' in \emph{IEDM Tech. Dig.}\hskip 1em plus 0.5em minus
  0.4em\relax IEEE, 2019, pp. 22--6. [Online]. Available:
  \url{10.1109/IEDM19573.2019.8993555}
\BIBentrySTDinterwordspacing

\bibitem{zhou2018new}
\BIBentryALTinterwordspacing
Z.~Zhou, P.~Huang, Y.~C. Xiang, W.~S. Shen, Y.~D. Zhao, Y.~L. Feng, B.~Gao,
  H.~Q. Wu, H.~Qian, L.~F. Liu, X.~Zhang, X.~Y. Liu, and J.~F. Kang, ``A new
  hardware implementation approach of bnns based on nonlinear 2t2r synaptic
  cell,'' in \emph{IEDM Tech. Dig.}\hskip 1em plus 0.5em minus 0.4em\relax
  IEEE, 2018, pp. 20--7. [Online]. Available: \url{10.1109/IEDM.2018.8614642}
\BIBentrySTDinterwordspacing

\bibitem{alibart2012high}
F.~Alibart, L.~Gao, B.~D. Hoskins, and D.~B. Strukov, ``High precision tuning
  of state for memristive devices by adaptable variation-tolerant algorithm,''
  \emph{Nanotechnology}, vol.~23, no.~7, p. 075201, 2012.

\bibitem{kvatinsky2015vteam}
S.~Kvatinsky, M.~Ramadan, E.~G. Friedman, and A.~Kolodny, ``Vteam: A general
  model for voltage-controlled memristors,'' \emph{IEEE Trans. Circuits Syst.
  II Express Briefs}, vol.~62, no.~8, pp. 786--790, 2015.

\bibitem{choi2014random}
S.~Choi, Y.~Yang, and W.~Lu, ``Random telegraph noise and resistance switching
  analysis of oxide based resistive memory,'' \emph{Nanoscale}, vol.~6, no.~1,
  pp. 400--404, 2014.

\bibitem{stathopoulos2019memristive}
S.~Stathopoulos, A.~Serb, A.~Khiat, M.~Ogorza{\l}ek, and T.~Prodromakis, ``A
  memristive switching uncertainty model,'' \emph{IEEE Transactions on Electron
  Devices}, vol.~66, no.~7, pp. 2946--2953, 2019.

\bibitem{paszke2017automatic}
\BIBentryALTinterwordspacing
A.~Paszke, S.~Gross, S.~Chintala, G.~Chanan, E.~Yang, Z.~DeVito, Z.~Lin,
  A.~Desmaison, L.~Antiga, and A.~Lerer, ``Automatic differentiation in
  pytorch,'' 2017. [Online]. Available:
  \url{https://openreview.net/forum?id=BJJsrmfCZ}
\BIBentrySTDinterwordspacing

\bibitem{hirtzlin2020digital}
\BIBentryALTinterwordspacing
T.~Hirtzlin, M.~Bocquet, B.~Penkovsky, J.-O. Klein, E.~Nowak, E.~Vianello,
  J.-M. Portal, and D.~Querlioz, ``Digital biologically plausible
  implementation of binarized neural networks with differential hafnium oxide
  resistive memory arrays,'' \emph{Frontiers in neuroscience}, vol.~13, p.
  1383, 2020. [Online]. Available: \url{10.3389/fnins.2019.01383}
\BIBentrySTDinterwordspacing

\bibitem{rusakov2020noisy}
\BIBentryALTinterwordspacing
D.~A. Rusakov, L.~P. Savtchenko, and P.~E. Latham, ``Noisy synaptic
  conductance: bug or a feature?'' \emph{Trends in Neurosciences}, 2020.
  [Online]. Available: \url{10.1016/j.tins.2020.03.009}
\BIBentrySTDinterwordspacing

\bibitem{raghavan2013microscopic}
\BIBentryALTinterwordspacing
N.~Raghavan, R.~Degraeve, A.~Fantini, L.~Goux, S.~Strangio, B.~Govoreanu,
  D.~Wouters, G.~Groeseneken, and M.~Jurczak, ``Microscopic origin of random
  telegraph noise fluctuations in aggressively scaled rram and its impact on
  read disturb variability,'' in \emph{Proc. IRPS}.\hskip 1em plus 0.5em minus
  0.4em\relax IEEE, 2013, pp. 5E--3. [Online]. Available:
  \url{10.1109/IRPS.2013.6532042}
\BIBentrySTDinterwordspacing

\bibitem{ambrogio2014impact}
\BIBentryALTinterwordspacing
S.~Ambrogio, S.~Balatti, V.~McCaffrey, D.~Wang, and D.~Ielmini, ``Impact of
  low-frequency noise on read distributions of resistive switching memory
  (rram),'' in \emph{IEDM Tech. Dig.}\hskip 1em plus 0.5em minus 0.4em\relax
  IEEE, 2014, pp. 14--4. [Online]. Available: \url{10.1109/IEDM.2014.7047051}
\BIBentrySTDinterwordspacing

\bibitem{ito2011modeling}
\BIBentryALTinterwordspacing
K.~Ito, T.~Matsumoto, S.~Nishizawa, H.~Sunagawa, K.~Kobayashi, and H.~Onodera,
  ``Modeling of random telegraph noise under circuit operation—simulation and
  measurement of rtn-induced delay fluctuation,'' in \emph{Proc. ISQED}.\hskip
  1em plus 0.5em minus 0.4em\relax IEEE, 2011, pp. 1--6. [Online]. Available:
  \url{10.1109/ISQED.2011.5770698}
\BIBentrySTDinterwordspacing

\bibitem{grasser2020noise}
\BIBentryALTinterwordspacing
T.~Grasser, \emph{Noise in Nanoscale Semiconductor Devices}.\hskip 1em plus
  0.5em minus 0.4em\relax Springer Nature, 2020. [Online]. Available:
  \url{10.1007/978-3-030-37500-3}
\BIBentrySTDinterwordspacing

\bibitem{kasdin1995discrete}
\BIBentryALTinterwordspacing
N.~J. Kasdin, ``Discrete simulation of colored noise and stochastic processes
  and 1/f/sup/spl alpha//power law noise generation,'' \emph{Proc. IEEE},
  vol.~83, no.~5, pp. 802--827, 1995. [Online]. Available:
  \url{10.1109/5.381848}
\BIBentrySTDinterwordspacing

\bibitem{ambrogio2018equivalent}
\BIBentryALTinterwordspacing
S.~Ambrogio, P.~Narayanan, H.~Tsai, R.~M. Shelby, I.~Boybat, C.~di~Nolfo,
  S.~Sidler, M.~Giordano, M.~Bodini, N.~C.~P. Farinha, B.~Killeen, C.~Cheng,
  Y.~Jaoudi, and G.~W. Burr, ``Equivalent-accuracy accelerated neural-network
  training using analogue memory,'' \emph{Nature}, vol. 558, no. 7708, p.~60,
  2018. [Online]. Available: \url{10.1038/s41586-018-0180-5}
\BIBentrySTDinterwordspacing

\bibitem{hubara2016binarized}
\BIBentryALTinterwordspacing
I.~Hubara, M.~Courbariaux, D.~Soudry, R.~El-Yaniv, and Y.~Bengio, ``Binarized
  neural networks,'' in \emph{Proc. NIPS}, 2016, pp. 4114--4122. [Online].
  Available: \url{https://arxiv.org/abs/1602.02830}
\BIBentrySTDinterwordspacing

\bibitem{rastegari2016xnor}
\BIBentryALTinterwordspacing
M.~Rastegari, V.~Ordonez, J.~Redmon, and A.~Farhadi, ``Xnor-net: Imagenet
  classification using binary convolutional neural networks,'' in \emph{Proc.
  ECCV}.\hskip 1em plus 0.5em minus 0.4em\relax Springer, 2016, pp. 525--542.
  [Online]. Available: \url{10.1007/978-3-319-46493-0\_32}
\BIBentrySTDinterwordspacing

\bibitem{bocquet2018memory}
M.~Bocquet, T.~Hirztlin, J.-O. Klein, E.~Nowak, E.~Vianello, J.-M. Portal, and
  D.~Querlioz, ``In-memory and error-immune differential rram implementation of
  binarized deep neural networks,'' in \emph{2018 IEEE International Electron
  Devices Meeting (IEDM)}.\hskip 1em plus 0.5em minus 0.4em\relax IEEE, 2018,
  pp. 20--6.

\bibitem{lastras2021ratio}
M.~A. Lastras-Monta{\~n}o, O.~Del Pozo-Zamudio, L.~Glebsky, M.~Zhao, H.~Wu, and
  K.-T. Cheng, ``Ratio-based multi-level resistive memory cells,'' \emph{Sci.
  Rep.}, vol.~11, no.~1, pp. 1--12, 2021.

\bibitem{zhao2014synchronous}
\BIBentryALTinterwordspacing
W.~Zhao, M.~Moreau, E.~Deng, Y.~Zhang, J.-M. Portal, J.-O. Klein, M.~Bocquet,
  H.~Aziza, D.~Deleruyelle, C.~Muller, D.~Querlioz, N.~Ben~Romdhane,
  D.~Ravelosona, and C.~Chappert, ``Synchronous non-volatile logic gate design
  based on resistive switching memories,'' \emph{{IEEE} Trans. Circuits Syst.
  {I}}, vol.~61, no.~2, pp. 443--454, 2014. [Online]. Available:
  \url{10.1109/TCSI.2013.2278332}
\BIBentrySTDinterwordspacing

\bibitem{kingma2015adam}
\BIBentryALTinterwordspacing
D.~Kingma and J.~Ba, ``Adam: A method for stochastic optimization,'' in
  \emph{Proc. ICLR}, 2015. [Online]. Available:
  \url{https://arxiv.org/abs/1412.6980}
\BIBentrySTDinterwordspacing

\bibitem{he2015delving}
\BIBentryALTinterwordspacing
K.~He, X.~Zhang, S.~Ren, and J.~Sun, ``Delving deep into rectifiers: Surpassing
  human-level performance on imagenet classification,'' in \emph{Proc. ICCV},
  2015, pp. 1026--1034. [Online]. Available: \url{10.1109/ICCV.2015.123}
\BIBentrySTDinterwordspacing

\bibitem{lecun2012efficient}
\BIBentryALTinterwordspacing
Y.~A. LeCun, L.~Bottou, G.~B. Orr, and K.-R. M{\"u}ller, ``Efficient
  backprop,'' in \emph{Neural networks: Tricks of the trade}.\hskip 1em plus
  0.5em minus 0.4em\relax Springer, 2012, pp. 9--48. [Online]. Available:
  \url{10.1007/978-3-642-35289-8\_3}
\BIBentrySTDinterwordspacing

\end{thebibliography}


\end{document}